\documentclass[11pt,a4paper]{article}

\usepackage[utf8]{inputenc}
\usepackage[T1]{fontenc}
\usepackage{amsmath,amssymb,amsfonts}
\usepackage{graphicx}
\usepackage{booktabs}
\usepackage{hyperref}
\usepackage[ruled,vlined,linesnumbered]{algorithm2e}
\usepackage{multirow}
\usepackage{caption}
\usepackage[margin=2.5cm]{geometry}
\usepackage{natbib}
\usepackage{xcolor}

\hypersetup{
  colorlinks=true,
  linkcolor=blue,
  citecolor=blue,
  urlcolor=blue
}

\title{Hybrid Metaheuristic Combining the Dragonfly Algorithm and Tabu Search for the Traveling Salesman Problem}

\author{
Bouketta Ammar\\
\small Ecole Nationale Superieure d'Informatique (ESI), Algiers, Algeria\\
\small \texttt{ja\_bouketta@esi.dz}
}

\date{}

\begin{document}
\maketitle

\begin{abstract}
The Traveling Salesman Problem (TSP) is a classical NP-hard combinatorial optimization problem that aims to find the shortest Hamiltonian cycle visiting each city exactly once and returning to the starting point. 
This paper proposes a hybrid metaheuristic for the TSP by combining the Dragonfly Algorithm (DA), a swarm-intelligence-based global search method, with Tabu Search (TS), a memory-based local search technique. 
The proposed method follows a High-Level Relay Hybridization (HRH) scheme, in which DA is first used to explore the solution space and generate a promising initial tour, while TS subsequently refines this solution through neighbourhood-based improvement and tabu memory. 
The hybrid approach is evaluated on standard TSPLIB benchmark instances, including \texttt{burma14}, \texttt{att48}, and \texttt{ch150}, and compared with standalone DA, standalone TS, and several classical metaheuristics such as Genetic Algorithm, Ant Colony Optimization, Particle Swarm Optimization, and Random Search. 
A systematic grid-search procedure is also conducted to study the influence of the main hyperparameters on solution quality and execution time. 
The experimental results indicate that the proposed hybrid can improve tour quality compared with the standalone DA and TS on the tested instances, highlighting the benefit of combining global exploration with local exploitation. 
However, the results also suggest that performance remains sensitive to parameter settings and problem size, motivating further validation on larger benchmarks and stronger TSP-specific baselines.
\end{abstract}

\section{Introduction}
\label{sec:intro}

The Traveling Salesman Problem (TSP) is one of the most classical and widely studied problems in combinatorial optimization. 
Given a set of cities and the pairwise distances between them, the objective is to find the shortest closed tour that visits each city exactly once and returns to the starting point. 
Although the problem is simple to state, it is computationally challenging: the number of feasible tours grows factorially with the number of cities, reaching $(n-1)!/2$ possible tours in the symmetric case. 
Karp~\citep{karp1972} showed that the closely related Hamiltonian-cycle problem is NP-complete, and the TSP is generally treated as an NP-hard optimization problem. 
As a result, exact methods become computationally expensive for large instances, motivating the development of heuristic and metaheuristic approaches capable of producing high-quality solutions within reasonable computation time.

A wide range of metaheuristics has been applied to the TSP, including Genetic Algorithms (GA)~\citep{holland1992}, Simulated Annealing (SA)~\citep{kirkpatrick1983}, Ant Colony Optimization (ACO)~\citep{dorigo1997}, and Particle Swarm Optimization (PSO)~\citep{kennedy1995}. 
These methods differ in the way they explore the search space and exploit promising regions. 
Population-based algorithms are generally effective at maintaining diversity and exploring multiple regions of the solution space, but they may require many iterations to converge. 
In contrast, trajectory-based methods can efficiently improve a single candidate solution through local search, but they are more vulnerable to premature convergence and entrapment in local optima. 
Hybrid metaheuristics have therefore become an important research direction, as they aim to combine complementary search behaviours and achieve a better balance between exploration and exploitation~\citep{talbi2002,blum2011}.

The Dragonfly Algorithm (DA), introduced by Mirjalili~\citep{mirjalili2016}, is a swarm-intelligence metaheuristic inspired by the static and dynamic swarming behaviours of dragonflies. 
Its search mechanism is based on separation, alignment, cohesion, attraction toward food sources, and distraction from enemies, allowing it to alternate between exploration and exploitation. 
DA has shown competitive performance on several optimization tasks~\citep{emambocus2021}; however, its direct application to discrete combinatorial problems such as the TSP remains challenging because the original algorithm is mainly defined in a continuous search space. 
On the other hand, Tabu Search (TS), proposed by Glover~\citep{glover1986}, is a well-established memory-based local search method that is particularly suitable for refining combinatorial solutions. 
By using a tabu list and aspiration criteria, TS can escape local optima while intensifying the search around promising solutions.

Motivated by the complementary characteristics of these two methods, this paper proposes a \emph{High-Level Relay Hybridization} (HRH) between DA and TS for the TSP. 
In the proposed approach, DA is first used as a global exploration mechanism to generate a promising initial tour. 
The best tour obtained by DA is then passed to TS, which performs a local refinement phase using neighbourhood-based search and tabu memory. 
This sequential hybridization aims to exploit the broad search capability of DA while benefiting from the intensification strength of TS.

The main contributions of this work are summarized as follows:
\begin{enumerate}
  \item We adapt the Dragonfly Algorithm to the discrete TSP domain using a permutation-based representation of candidate tours.
  \item We propose a High-Level Relay Hybridization scheme in which DA performs global exploration and TS performs local refinement.
  \item We conduct a systematic hyperparameter study to analyse the influence of key DA and TS parameters on solution quality and execution time.
  \item We evaluate the proposed hybrid method on standard TSPLIB benchmark instances of increasing size and compare it with standalone DA, standalone TS, and several classical metaheuristics.
\end{enumerate}

The remainder of the paper is organized as follows. 
Section~\ref{sec:background} reviews the TSP, relevant metaheuristics, and hybridization strategies. 
Section~\ref{sec:method} presents the proposed DA--TS hybrid algorithm. 
Section~\ref{sec:experiments} describes the experimental setup and discusses the results. 
Finally, Section~\ref{sec:conclusion} concludes the paper and outlines future research directions.
\section{Background}
\label{sec:background}
\subsection{The Traveling Salesman Problem}
\label{subsec:tsp}

The Traveling Salesman Problem (TSP) is a classical combinatorial optimization problem in which a salesman must visit a set of cities exactly once and return to the starting city while minimizing the total travel cost. 
It can be modelled as a complete weighted graph $G = (V,E)$, where $V = \{1,2,\ldots,n\}$ is the set of cities and $E$ is the set of edges connecting every pair of cities. 
Each edge $(i,j)$ is associated with a non-negative weight $d_{ij}$, which represents the distance or travel cost between city $i$ and city $j$.

\begin{figure*}[t]
    \centering
    \includegraphics[width=0.95\textwidth]{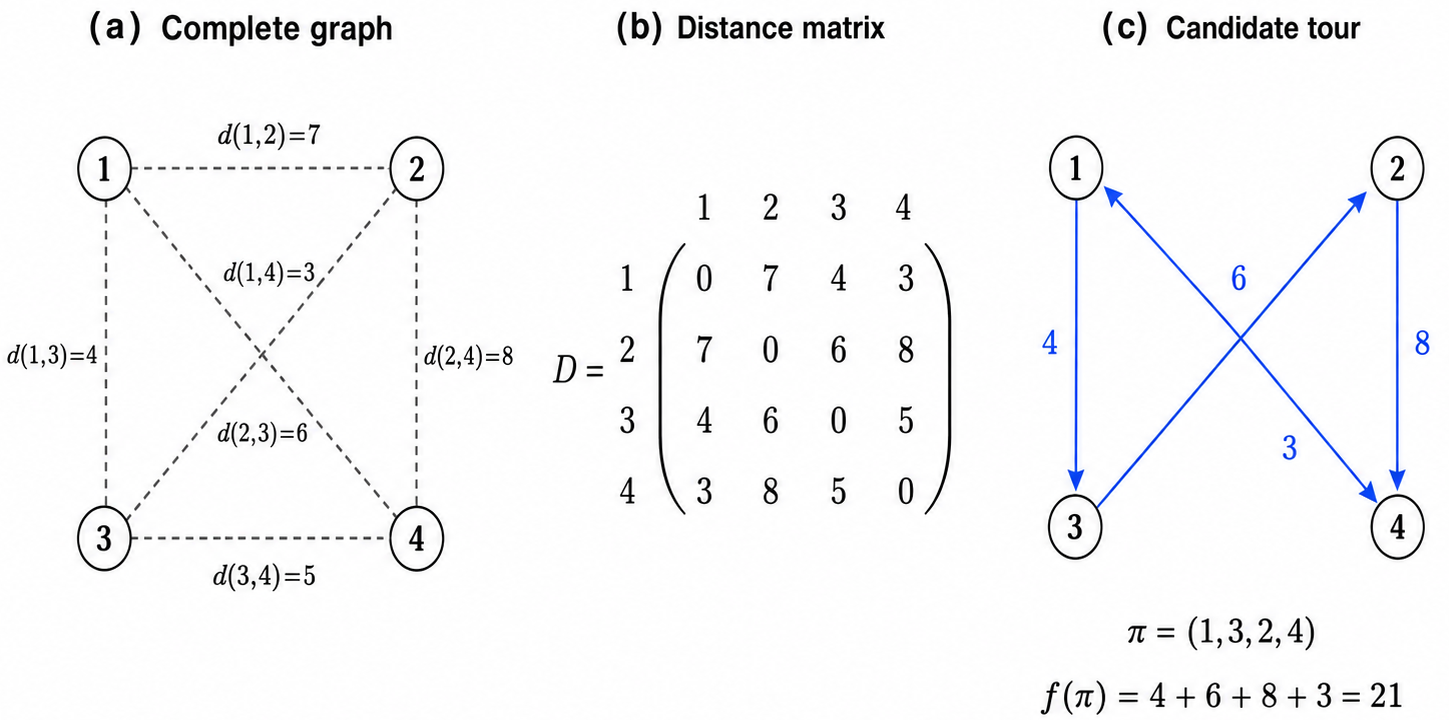}
    \caption{Illustration of the Traveling Salesman Problem. 
    (a) A complete weighted graph represents the cities and all pairwise travel costs. 
    (b) The same information can be stored in a symmetric distance matrix. 
    (c) A candidate tour is represented as a permutation of cities and its cost is obtained by summing the selected edge weights.}
    \label{fig:tsp_illustration}
\end{figure*}

The input data of a TSP instance may be provided either as city coordinates or directly as a distance matrix. 
When coordinates are given, the pairwise distances are first computed to construct a distance matrix $D = (d_{ij})$, where each entry $d_{ij}$ denotes the travel cost between city $i$ and city $j$. 
Figure~\ref{fig:tsp_illustration} illustrates this representation on a small example with four cities: the complete graph defines all pairwise distances, the corresponding distance matrix stores these values, and a candidate tour is evaluated by summing the distances of the selected edges.

A candidate solution to the TSP is represented as a permutation 
$\pi = (\pi(1), \pi(2), \ldots, \pi(n))$ of the $n$ cities. 
This permutation defines the visiting order of the tour:
\[
\pi(1) \rightarrow \pi(2) \rightarrow \cdots \rightarrow \pi(n) \rightarrow \pi(1).
\]
The objective is to find the permutation that minimizes the total tour cost:
\begin{equation}
  \label{eq:tsp}
  f(\pi) = \sum_{k=1}^{n-1} d_{\pi(k),\,\pi(k+1)} + d_{\pi(n),\,\pi(1)}.
\end{equation}
The first term in Eq.~\eqref{eq:tsp} sums the distances between consecutive cities in the tour, while the second term adds the return distance from the last visited city back to the starting city.

For example, in Figure~\ref{fig:tsp_illustration}(c), the candidate tour is $\pi=(1,3,2,4)$. 
Its total cost is computed as:
\[
f(\pi) = d_{1,3} + d_{3,2} + d_{2,4} + d_{4,1}
       = 4 + 6 + 8 + 3 = 21.
\]
This example shows that evaluating a TSP solution consists simply of reading the corresponding edge weights from the distance matrix and summing them along the complete closed tour.

When the distance matrix is symmetric, i.e., $d_{ij} = d_{ji}$ for all pairs of cities, the problem is called the \emph{symmetric TSP}. 
Otherwise, it is referred to as the \emph{asymmetric TSP}. 
In this work, we focus on the symmetric TSP, which is the setting used by the benchmark instances considered in our experiments.

The TSP has many practical applications, including logistics, vehicle routing, circuit board drilling, genome sequencing, and path planning. 
Although the problem is easy to formulate, it is computationally difficult because the number of feasible tours grows factorially with the number of cities. 
For the symmetric TSP, there are $(n-1)!/2$ distinct tours. 
As a result, exact methods such as branch-and-bound~\citep{little1963} and cutting-plane algorithms can become computationally expensive for large instances. 
Approximation algorithms, such as the Christofides--Serdyukov algorithm~\citep{christofides1976}, provide theoretical guarantees for metric instances, but in practice, metaheuristic approaches are widely used because they offer a flexible trade-off between solution quality and computation time.

\subsection{Metaheuristic Approaches to the TSP}
\label{subsec:metaheuristics}

Because the Traveling Salesman Problem (TSP) is computationally difficult to solve exactly for large instances, a wide range of heuristic and metaheuristic methods has been developed to obtain high-quality approximate solutions within reasonable computation time. 
These methods can be broadly grouped into constructive heuristics, improvement heuristics, population-based metaheuristics, and trajectory-based metaheuristics.

\paragraph{Constructive heuristics.}
Constructive heuristics build a feasible tour step by step, starting from an empty or partial solution. 
The Nearest Neighbour (NN) heuristic is one of the simplest examples: starting from an initial city, it repeatedly visits the closest unvisited city until all cities have been included in the tour. 
The Cheapest Insertion heuristic follows a different strategy by inserting, at each step, the city that causes the smallest increase in the current tour length. 
These methods are computationally efficient, often running in $O(n^2)$ time, but they are greedy and do not provide guarantees on solution quality. 
They are therefore commonly used to generate initial solutions for more advanced improvement or metaheuristic methods.

\paragraph{Improvement heuristics.}
Improvement heuristics start from an existing feasible tour and iteratively modify it in order to reduce its total cost. 
The 2-opt heuristic~\citep{croes1958} is a classical local search method that removes two edges from a tour and reconnects the resulting paths whenever this produces a shorter tour. 
More advanced variants, such as 3-opt and the Lin--Kernighan heuristic~\citep{lin1973}, explore larger neighbourhoods and are generally more effective, although they require higher computational effort. 
Such methods remain important in modern TSP solvers because they provide strong local refinement and are frequently embedded inside hybrid metaheuristics.

\paragraph{Population-based metaheuristics.}
Population-based metaheuristics maintain several candidate solutions simultaneously and use collective search mechanisms to explore different regions of the solution space. 
Genetic Algorithms (GAs) evolve a population of tours through selection, crossover, and mutation operators~\citep{holland1992}. 
Ant Colony Optimization (ACO) constructs tours probabilistically by using artificial pheromone trails that are reinforced according to the quality of previously found solutions~\citep{dorigo1997}. 
Particle Swarm Optimization (PSO), originally introduced for continuous optimization~\citep{kennedy1995}, has also been adapted to the discrete TSP by redefining particle positions and movements in terms of permutations or swap-based operations. 
These approaches are effective for global exploration, but they may require careful parameter tuning and can converge slowly if not combined with local improvement mechanisms.

\paragraph{Trajectory-based metaheuristics.}
Trajectory-based metaheuristics improve a single candidate solution over time by moving through its neighbourhood. 
Simulated Annealing (SA) accepts not only improving moves but also, with a decreasing probability, some worsening moves, which allows the search to escape local optima~\citep{kirkpatrick1983}. 
Tabu Search (TS) uses short-term memory, typically in the form of a tabu list, to prevent cycling and discourage recently visited moves~\citep{glover1986,glover1997}. 
By combining intensification around promising solutions with diversification mechanisms, TS has been widely used for combinatorial optimization problems such as the TSP.

\paragraph{Hybrid metaheuristics.}
Recent research has increasingly focused on hybrid metaheuristics, where complementary search mechanisms are combined to improve robustness and solution quality~\citep{talbi2002,blum2011}. 
For the TSP and its variants, hybrid methods often combine a population-based exploration phase with a local search or memory-based refinement phase. 
This design is motivated by the observation that population-based algorithms are generally effective at discovering promising regions of the search space, while local search methods are more efficient at intensifying the search around high-quality candidate tours. 
The proposed Dragonfly Algorithm--Tabu Search hybrid follows this principle: the Dragonfly Algorithm (DA) is used to explore the space of candidate tours, and TS is then applied to refine the best solution found by DA.

\subsection{Hybridization of Metaheuristics}
\label{subsec:hybrid}

Hybridization is a widely used strategy in metaheuristic optimization, particularly for difficult combinatorial problems such as the Traveling Salesman Problem (TSP). 
The main idea is to combine two or more search mechanisms in order to exploit their complementary strengths. 
For example, a population-based method can provide broad exploration of the search space, while a local search method can intensify the search around promising candidate solutions. 
This combination is often more effective than relying on a single metaheuristic, especially when the problem requires both diversification and accurate local refinement.

Talbi~\citep{talbi2002} proposed a widely adopted taxonomy for classifying hybrid metaheuristics. 
At the design level, hybrid methods can be classified as either \emph{low-level hybridization} (LLH) or \emph{high-level hybridization} (HLH). 
In LLH, internal components of one algorithm are embedded into another algorithm, such as using a local search operator inside a genetic algorithm. 
In contrast, HLH combines complete metaheuristics while preserving their individual structures. 
The interaction between the combined algorithms can further be categorized as either \emph{relay} or \emph{teamwork}. 
In relay hybridization, algorithms are executed sequentially, where the output of one algorithm becomes the input of another. 
In teamwork hybridization, the algorithms cooperate during the search, often by exchanging information in parallel or iteratively. 
Therefore, a \emph{High-Level Relay Hybridization} (HRH) scheme refers to a sequential combination of complete algorithms, where the first method produces a solution that is then refined or improved by the second method.

Several hybrid metaheuristics have been proposed for the TSP and its variants. 
Early examples include Genetic Algorithm (GA) combined with 2-opt local search~\citep{ulder1991} and memetic algorithms that integrate evolutionary search with local improvement mechanisms~\citep{moscato1989}. 
Other studies have combined Ant Colony Optimization (ACO) with Simulated Annealing (SA), Tabu Search (TS), or other local search procedures to improve the balance between exploration and exploitation~\citep{wang2015}. 
More recent works continue to show that hybridization remains relevant for routing problems and TSP variants, including combinations of GA and ACO for the Traveling Salesman Problem with Drone (TSP-D), hybrid algorithms for generalized TSP variants, and recent surveys on heuristic and metaheuristic methods for TSP-related problems. 

The common motivation behind these approaches is that no single metaheuristic is universally dominant across all TSP instances. 
Population-based methods are usually effective for global exploration and maintaining solution diversity, but they may converge slowly or require many evaluations. 
Trajectory-based methods, such as TS, are more effective for local intensification, but their final performance strongly depends on the quality of the initial solution. 
This observation motivates the hybrid strategy adopted in this work. 
The proposed method follows an HRH scheme in which the Dragonfly Algorithm (DA) first explores the permutation search space and identifies a promising tour. 
The best tour found by DA is then passed to TS, which performs neighbourhood-based refinement using tabu memory. 
This design aims to combine the exploration ability of DA with the exploitation strength of TS.

\subsection{The Dragonfly Algorithm}
\label{subsec:da}

The DA~\citep{mirjalili2016} models five behaviours observed in dragonfly swarms:

\begin{enumerate}
  \item \textbf{Separation} ($S_i$): avoiding collisions with neighbours.
  \item \textbf{Alignment} ($A_i$): matching velocity with neighbours.
  \item \textbf{Cohesion} ($C_i$): moving toward the neighbourhood centre of mass.
  \item \textbf{Attraction to food} ($F_i$): steering toward the best-known solution (food source).
  \item \textbf{Distraction from enemy} ($E_i$): fleeing from the worst-known solution (enemy).
\end{enumerate}

These factors are computed as follows. Let $X_i$ denote the position of dragonfly $i$, $X_j$ ($j = 1, \ldots, N$) the positions of its $N$ neighbours, and $V_j$ their velocities:
\begin{align}
  S_i &= -\sum_{j=1}^{N} (X_i - X_j), \label{eq:sep}\\
  A_i &= \frac{1}{N}\sum_{j=1}^{N} V_j, \label{eq:align}\\
  C_i &= \frac{1}{N}\sum_{j=1}^{N} X_j - X_i, \label{eq:coh}\\
  F_i &= X^{+} - X_i, \label{eq:food}\\
  E_i &= X^{-} + X_i, \label{eq:enemy}
\end{align}
where $X^{+}$ is the food-source position (best solution) and $X^{-}$ is the enemy position (worst solution).

The step vector and position vector are updated as:
\begin{align}
  \Delta X_i^{t+1} &= \bigl(s\,S_i + a\,A_i + c\,C_i + f\,F_i + e\,E_i\bigr) + w\,\Delta X_i^{t}, \label{eq:step}\\
  X_i^{t+1} &= X_i^{t} + \Delta X_i^{t+1}, \label{eq:pos}
\end{align}
where $s, a, c, f, e$ are weighting coefficients for the five factors and $w$ is an inertia weight.
If a dragonfly has no neighbours, its position is updated via a L\'evy flight:
\begin{equation}
  X_i^{t+1} = X_i^{t} + \mathrm{L\acute{e}vy}(d) \times X_i^{t}, \label{eq:levy}
\end{equation}
where $d$ is the dimensionality of the search space.
The neighbourhood radius is increased proportionally with the iteration count, gradually transitioning the swarm from exploration (static swarming) to exploitation (dynamic swarming).

\subsection{Tabu Search}
\label{subsec:ts}

Tabu Search~\citep{glover1986,glover1997} is a trajectory-based metaheuristic that enhances local search by incorporating memory.
Its key components in the TSP context are as follows.

\begin{itemize}
  \item \textbf{Initial solution}: typically generated by a constructive heuristic (e.g., nearest neighbour).
  \item \textbf{Neighbourhood}: defined by a move operator (e.g., 2-opt swap); at each iteration, the best non-tabu neighbour is selected.
  \item \textbf{Tabu list}: a short-term memory that records recently performed moves (e.g., swapped city pairs) and forbids their reversal for a fixed number of iterations.
  \item \textbf{Aspiration criterion}: overrides the tabu status of a move if it produces a solution better than the current best.
  \item \textbf{Diversification}: long-term mechanisms that redirect the search to unexplored regions when stagnation is detected.
\end{itemize}

\section{Proposed Hybrid Algorithm}
\label{sec:method}

This section presents the proposed hybrid algorithm for solving the Traveling Salesman Problem (TSP). 
The method combines the Dragonfly Algorithm (DA) and Tabu Search (TS) through a High-Level Relay Hybridization (HRH) strategy. 
The main idea is to use DA as a global exploration mechanism to identify a promising tour, and then use TS as a local exploitation mechanism to refine this tour and improve the final solution quality.

\subsection{Overall Architecture}
\label{subsec:arch}

\begin{figure}[htpb]
  \centering
  \includegraphics[width=0.4\linewidth]{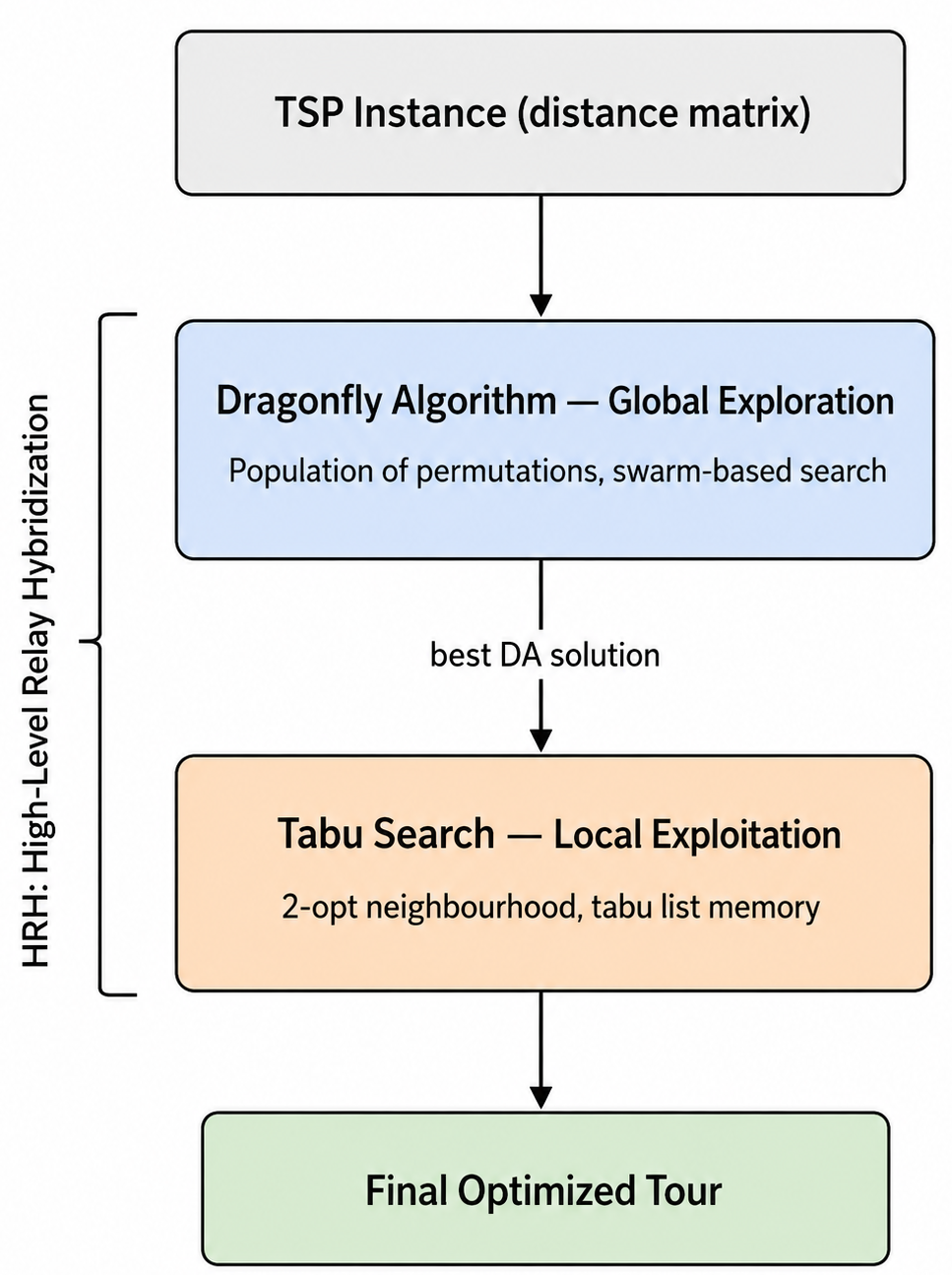}
  \caption{Overall architecture of the proposed High-Level Relay Hybridization (HRH) scheme. 
  The Dragonfly Algorithm performs global exploration over a population of TSP permutations, while Tabu Search locally refines the best solution found by DA.}
  \label{fig:architecture}
\end{figure}

The proposed method follows a sequential HRH architecture, as illustrated in Figure~\ref{fig:architecture}. 
The input is a TSP instance represented by a distance matrix. 
First, the DA phase explores the search space using a population of candidate tours encoded as permutations. 
After a fixed number of DA iterations, the best solution found by the population is selected and passed to the TS phase. 
Then, TS starts from this solution and applies neighbourhood-based local search using 2-opt moves, tabu memory, and an aspiration criterion. 
The final output of the hybrid method is the best tour obtained after the TS refinement phase.

This architecture is high-level because DA and TS remain two independent metaheuristics, and it is relay-based because the output of the first algorithm is used as the input of the second one. 
This design aims to combine the exploration capability of DA with the exploitation strength of TS.

\subsection{Solution Encoding}
\label{subsec:encoding}

A TSP solution is encoded as a permutation 
$\pi = (\pi_1, \pi_2, \ldots, \pi_n)$ of the city indices $\{0,1,\ldots,n-1\}$. 
The tour visits the cities in the order
\[
\pi_1 \rightarrow \pi_2 \rightarrow \cdots \rightarrow \pi_n \rightarrow \pi_1.
\]
For example, the permutation $(0,2,4,1,3)$ represents the closed tour
\[
0 \rightarrow 2 \rightarrow 4 \rightarrow 1 \rightarrow 3 \rightarrow 0.
\]
This encoding is suitable for the TSP because each city appears exactly once in the permutation, ensuring that every candidate solution represents a valid tour.

\subsection{Dragonfly Phase}
\label{subsec:da_phase}

The first phase of the proposed algorithm is based on DA. 
In the original formulation, DA operates in a continuous search space. 
However, the TSP is a discrete combinatorial problem where candidate solutions are permutations of cities. 
Therefore, in this work, each dragonfly position is represented as a valid TSP permutation rather than a continuous vector.

The DA behaviours (separation, alignment, cohesion, food attraction, and enemy distraction) are used to guide the search process. 
Instead of directly updating continuous positions, the movement of a dragonfly is implemented through permutation-preserving operators, mainly swap-based operations. 
These operators modify the order of cities while keeping the solution valid. 
When a dragonfly has neighbouring individuals, the update is guided by the DA movement equations and translated into swap operations applied to the current permutation. 
When no neighbouring dragonfly is available, the Lévy-flight mechanism is implemented as a sequence of random swaps, which increases diversity and helps the algorithm explore new regions of the search space.

The neighbourhood radius increases progressively with the iteration counter. 
At the beginning of the search, a smaller radius encourages exploration by allowing dragonflies to move more independently. 
As the iterations progress, the radius increases, promoting convergence toward promising regions. 
At the end of the DA phase, the best permutation found across the population is retained and used as the initial solution for the TS phase.

\subsection{Tabu Search Phase}
\label{subsec:ts_phase}

The second phase applies TS to refine the best solution produced by DA. 
Starting from this initial tour, TS explores the neighbourhood of the current solution using the 2-opt operator. 
A 2-opt move removes two edges from the tour and reconnects the resulting segments by reversing part of the route. 
This operation is widely used in TSP local search because it can efficiently reduce crossing edges and improve tour quality.

At each TS iteration, candidate neighbours are generated and evaluated. 
The best admissible neighbour is selected as the next current solution. 
A move is considered admissible if it is not present in the tabu list, or if it satisfies the aspiration criterion by improving the best solution found so far. 
The tabu list stores recently applied moves for a fixed number of iterations in order to avoid cycling and encourage diversification. 
The TS phase terminates after a predefined number of iterations, and the best solution found during this phase is returned as the final output of the hybrid algorithm.

\subsection{Execution Workflow}
\label{subsec:workflow}

\begin{figure}[htpb]
  \centering
  \includegraphics[width=0.7\textwidth]{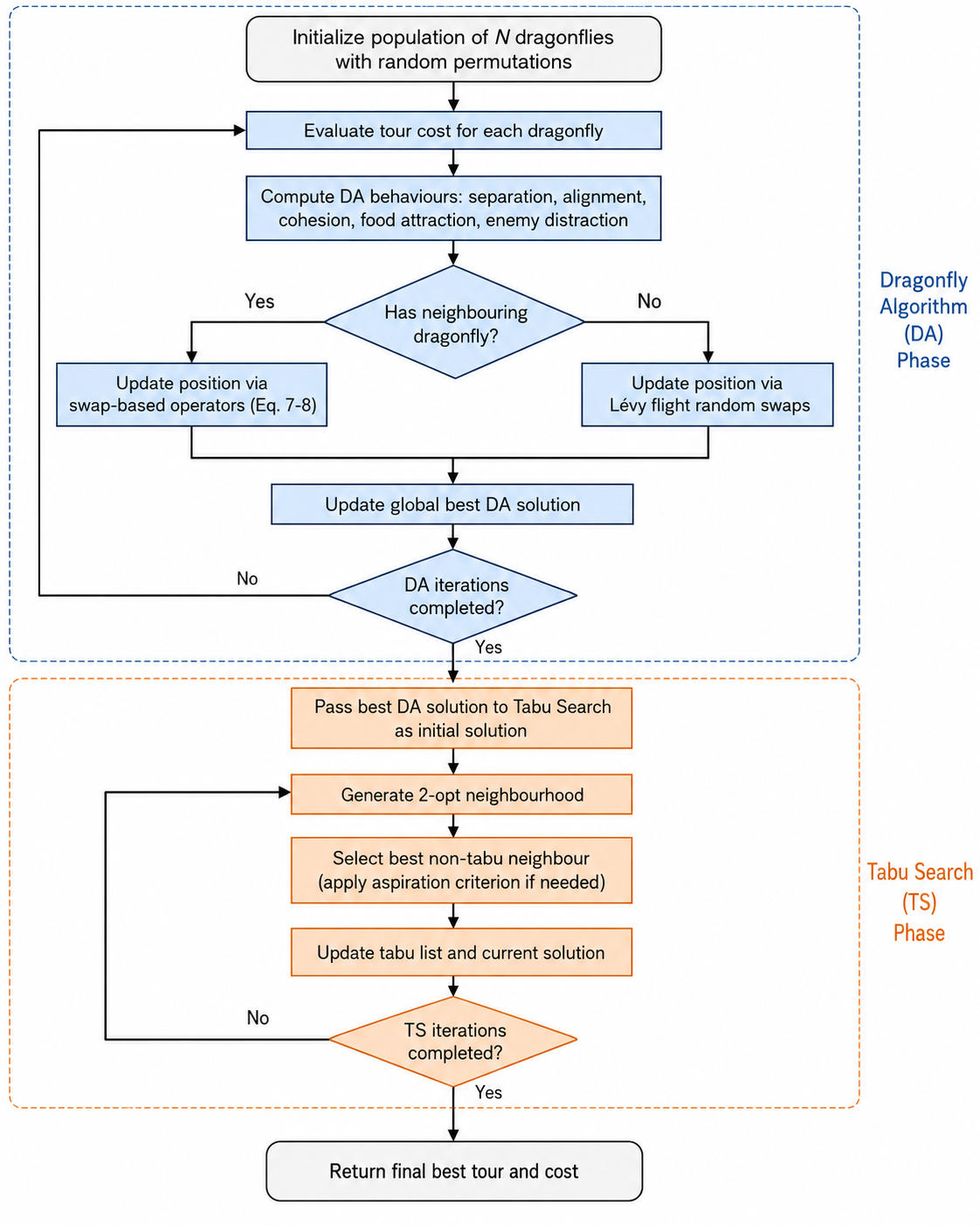}
  \caption{Detailed execution workflow of the proposed Dragonfly Algorithm--Tabu Search hybrid method. 
  The upper part corresponds to the Dragonfly Algorithm phase, which performs population-based global exploration. 
  The lower part corresponds to the Tabu Search phase, which refines the best DA solution using 2-opt neighbourhood search and tabu memory.}
  \label{fig:workflow}
\end{figure}

Figure~\ref{fig:workflow} provides the detailed execution workflow of the proposed DA--TS hybrid method. 
The algorithm starts by generating a population of random TSP permutations and evaluating their tour costs. 
Then, DA iteratively updates the population using either neighbour-guided swap-based operations or Lévy-flight random swaps. 
After the DA stopping criterion is satisfied, the best DA solution is transferred to TS. 
The TS phase then repeatedly generates 2-opt neighbours, selects the best non-tabu candidate, updates the tabu list, and keeps track of the best solution found. 
The complete process ends by returning the final best tour and its corresponding cost.

\subsection{Pseudocode}
\label{subsec:pseudo}

Algorithm~\ref{alg:hybrid} summarises the complete hybrid procedure. 
The first phase initializes and updates the DA population until the DA stopping criterion is reached. 
The second phase initializes TS with the best DA solution and improves it through repeated 2-opt neighbourhood search.

\begin{algorithm}[htpb]
\SetAlgoLined
\KwIn{Distance matrix $D$, DA parameters ($s, a, c, f, e, w, d$, population size $P$, iterations $T_{\mathrm{DA}}$), TS parameters (tabu list size $L$, iterations $T_{\mathrm{TS}}$)}
\KwOut{Best tour $\pi^*$ and its cost $f(\pi^*)$}
\BlankLine

\textbf{Phase 1: Dragonfly Algorithm}\\
Initialise population $\{\pi_1, \ldots, \pi_P\}$ with random valid permutations\;
Evaluate the tour cost $f(\pi_i)$ for each dragonfly\;
$\pi^* \leftarrow \arg\min_i f(\pi_i)$\;

\For{$t = 1$ \KwTo $T_{\mathrm{DA}}$}{
  Update the neighbourhood radius according to the iteration index\;
  \For{each dragonfly $i = 1, \ldots, P$}{
    Compute the DA behavioural factors $S_i$, $A_i$, $C_i$, $F_i$, and $E_i$\;
    \eIf{dragonfly $i$ has neighbouring dragonflies}{
      Update $\pi_i$ using swap-based operators guided by the DA update rule\;
    }{
      Update $\pi_i$ using Lévy-flight-inspired random swaps\;
    }
    Evaluate $f(\pi_i)$\;
    \If{$f(\pi_i) < f(\pi^*)$}{
      $\pi^* \leftarrow \pi_i$\;
    }
  }
}

\BlankLine
\textbf{Phase 2: Tabu Search}\\
$\pi_{\mathrm{current}} \leftarrow \pi^*$\;
Initialise tabu list $\mathcal{T} \leftarrow \emptyset$\;

\For{$t = 1$ \KwTo $T_{\mathrm{TS}}$}{
  Generate the 2-opt neighbourhood $\mathcal{N}(\pi_{\mathrm{current}})$\;
  Select the best candidate $\pi' \in \mathcal{N}(\pi_{\mathrm{current}})$ such that the corresponding move is not tabu, or satisfies the aspiration criterion\;
  $\pi_{\mathrm{current}} \leftarrow \pi'$\;
  Update the tabu list $\mathcal{T}$\;
  \If{$f(\pi') < f(\pi^*)$}{
    $\pi^* \leftarrow \pi'$\;
  }
}

\Return $\pi^*, f(\pi^*)$\;
\caption{Hybrid Dragonfly--Tabu Search algorithm for the TSP}
\label{alg:hybrid}
\end{algorithm}

\section{Experiments and Results}
\label{sec:experiments}

This section evaluates the proposed Dragonfly Algorithm--Tabu Search (DA--TS) hybrid method on standard Traveling Salesman Problem (TSP) benchmark instances. 
The objective is to assess whether the proposed High-Level Relay Hybridization (HRH) improves solution quality compared with standalone DA, standalone TS, and other classical heuristic and metaheuristic baselines. 

\subsection{Benchmark Instances}
\label{subsec:benchmarks}

We evaluate the proposed algorithm on three standard instances from the TSPLIB library~\citep{reinelt1991}, selected to cover increasing problem sizes:
\begin{itemize}
  \item \textbf{\texttt{burma14}}: 14 cities in Myanmar, with known optimum $3\,323$.
  \item \textbf{\texttt{att48}}: 48 state capitals in the United States, with known optimum $10\,628$.
  \item \textbf{\texttt{ch150}}: 150 cities, with known optimum $6\,528$.
\end{itemize}

Each instance is converted into a distance matrix $D=(d_{ij})$, where $d_{ij}$ denotes the travel cost between cities $i$ and $j$. 
The distance computation follows the edge-weight type specified in each TSPLIB instance, such as geographical distances for \texttt{burma14}, pseudo-Euclidean distances for \texttt{att48}, and Euclidean distances for \texttt{ch150}. 
The objective value of a candidate tour is then computed according to Eq.~\eqref{eq:tsp}.

To quantify solution quality, we report the relative gap to the known optimum:
\begin{equation}
\label{eq:gap}
\mathrm{Gap}(\%) =
\frac{f_{\mathrm{best}} - f^{*}}{f^{*}} \times 100,
\end{equation}
where $f_{\mathrm{best}}$ is the best tour cost obtained by the algorithm and $f^{*}$ is the known optimal cost.

\subsection{Execution Environment and Protocol}
\label{subsec:env}

All experiments were conducted on a machine equipped with an Intel Core i5-8250U CPU and 16\,GB of RAM, running Ubuntu 22.04 with Python~3.10. 
All algorithms were implemented in the same programming environment and executed on the same machine to ensure a fair comparison of execution time.

Since DA, Genetic Algorithm (GA), Ant Colony Optimization (ACO), Particle Swarm Optimization (PSO), Random Search (RS), and the proposed DA--TS hybrid are stochastic methods, each algorithm was executed 10 times per instance using different random seeds. 
For each method, we report the best cost found across the runs and the average execution time. 
In the final version, it is recommended to also report the mean cost and standard deviation over the runs to better assess robustness.

\subsection{Hyperparameter Tuning}
\label{subsec:tuning}

The proposed hybrid algorithm involves hyperparameters related to both the DA exploration phase and the TS refinement phase. 
The DA parameters include the number of iterations $T_{\mathrm{DA}}$, the population size $P$, the behavioural weights for separation $s$, alignment $a$, cohesion $c$, food attraction $f$, enemy distraction $e$, the inertia weight $w$, and the neighbourhood-radius update parameter $d$. 
The TS parameters include the number of TS iterations $T_{\mathrm{TS}}$ and the tabu list size $L$.

A grid search was performed on the \texttt{att48} instance by varying one parameter at a time while keeping the others fixed. 
Although this strategy does not guarantee a globally optimal configuration, it provides useful insight into the effect of each parameter on solution quality and execution time. 
Table~\ref{tab:param_effects} summarizes the main qualitative observations.

The final configuration used in the following experiments was:
\[
T_{\mathrm{DA}} = 200,\quad
P = 100,\quad
s = 0.5,\quad
a = 0.3,\quad
c = 0.4,\quad
f = 0.7,\quad
e = 0.4,\quad
w = 0.5,\quad
d = 100,\quad
T_{\mathrm{TS}} = 500,\quad
L = 20.
\]

\begin{table}[ht]
\centering
\caption{Summary of hyperparameter effects observed during grid search on \texttt{att48}.}
\label{tab:param_effects}
\begin{tabular}{@{}lp{8.5cm}@{}}
\toprule
\textbf{Parameter} & \textbf{Observation} \\
\midrule
$T_{\mathrm{DA}}$ & Increasing DA iterations improves the cost during early search, with diminishing returns after approximately 200 iterations. \\
$P$ & Larger populations improve exploration up to approximately 100 individuals, after which the gain becomes limited compared with the additional runtime. \\
$d$ & Very small radius updates cause premature convergence, while moderate values provide a better exploration--exploitation balance. \\
$c$ & Higher cohesion accelerates convergence but may increase the risk of local-optimum trapping. \\
$s$ & Separation promotes diversity; moderate values around $0.4$--$0.6$ give the most stable behaviour. \\
$e$ & Moderate enemy distraction encourages exploration without excessively disrupting convergence. \\
$f$ & Stronger food attraction guides the swarm toward the best-known solution and improves intensification. \\
$T_{\mathrm{TS}}$ & Increasing TS iterations improves refinement up to approximately 500 iterations, with limited gains afterwards. \\
$L$ & Small tabu lists may cause cycling, while very large lists may over-restrict the search. Values around 15--25 provide a good compromise. \\
\bottomrule
\end{tabular}
\end{table}

\subsection{Standalone DA, Standalone TS, and Hybrid DA--TS}
\label{subsec:standalone}

Table~\ref{tab:standalone} compares standalone DA, standalone TS, and the proposed DA--TS hybrid on the three TSPLIB instances. 
For each method, we report the best tour cost, the gap to the known optimum, and the average runtime over 10 runs.

\begin{table*}[t]
\centering
\caption{Performance comparison between standalone DA, standalone TS, and the proposed DA--TS hybrid.}
\label{tab:standalone}
\begin{tabular}{@{}l rrr rrr rrr@{}}
\toprule
& \multicolumn{3}{c}{\textbf{\texttt{burma14}} ($f^{*}=3\,323$)} 
& \multicolumn{3}{c}{\textbf{\texttt{att48}} ($f^{*}=10\,628$)}
& \multicolumn{3}{c}{\textbf{\texttt{ch150}} ($f^{*}=6\,528$)} \\
\cmidrule(lr){2-4} \cmidrule(lr){5-7} \cmidrule(lr){8-10}
\textbf{Algorithm} 
& Cost & Gap & Time 
& Cost & Gap & Time 
& Cost & Gap & Time \\
\midrule
DA              
& 3\,695 & 11.2\% & 2.4\,s  
& 13\,842 & 30.2\% & 18.3\,s  
& 14\,520 & 122.4\% & 42.7\,s \\
TS              
& 3\,371 & 1.4\%  & 0.08\,s 
& 12\,286 & 15.6\% & 0.42\,s  
& 8\,194  & 25.5\%  & 3.8\,s  \\
\textbf{DA--TS Hybrid} 
& \textbf{3\,323} & \textbf{0.0\%} & 2.5\,s 
& \textbf{11\,352} & \textbf{6.8\%} & 19.6\,s 
& \textbf{7\,482} & \textbf{14.6\%} & 48.1\,s \\
\bottomrule
\end{tabular}
\end{table*}

The hybrid obtains the best result among the three compared methods on all instances. 
On \texttt{burma14}, it reaches the known optimum of $3\,323$, while standalone TS remains slightly above the optimum and standalone DA has a larger gap. 
On \texttt{att48}, the hybrid reduces the optimality gap to $6.8\%$, compared with $15.6\%$ for TS and $30.2\%$ for DA. 
On \texttt{ch150}, the hybrid also improves over both standalone components, reducing the gap from $25.5\%$ for TS to $14.6\%$.

These results support the motivation behind the proposed HRH strategy. 
DA provides a population-based exploration phase that can identify promising regions of the search space, while TS refines the best DA solution using memory-guided local search. 
The improvement over standalone DA is particularly large, suggesting that DA alone has difficulty converging efficiently in the discrete TSP search space. 
The improvement over standalone TS is smaller but consistent, indicating that the DA phase provides a better starting point for TS.

\subsection{Comparison with Additional Baselines}
\label{subsec:comparison}

To position the proposed hybrid method against established approaches, we compare it with four additional metaheuristics: GA, ACO, PSO, and RS. 
We also include three classical heuristic baselines: Nearest Neighbour (NN), Farthest Insertion (FI), and 2-opt initialized from NN. 
All metaheuristics were implemented in Python under the same execution environment and assigned comparable computational budgets where possible. 
The constructive heuristics are included mainly as fast reference baselines.

Table~\ref{tab:comparison_full} reports the comparison on the three TSPLIB instances.

\begin{table*}[t]
\centering
\caption{Comparison of the proposed hybrid with heuristic and metaheuristic baselines.}
\label{tab:comparison_full}
\begin{tabular}{@{}l rr rr rr@{}}
\toprule
& \multicolumn{2}{c}{\textbf{\texttt{burma14}}} 
& \multicolumn{2}{c}{\textbf{\texttt{att48}}} 
& \multicolumn{2}{c}{\textbf{\texttt{ch150}}} \\
\cmidrule(lr){2-3} \cmidrule(lr){4-5} \cmidrule(lr){6-7}
\textbf{Algorithm} & Cost & Time (s) & Cost & Time (s) & Cost & Time (s) \\
\midrule
\multicolumn{7}{@{}l}{\textit{Constructive heuristics and local search}} \\
\quad NN (best of multi-start) & 3\,843 & $<$0.01 & 12\,861 & $<$0.01 & 8\,344 & 0.02 \\
\quad Farthest Insertion       & 4\,210 & $<$0.01 & 13\,948 & $<$0.01 & 9\,127 & 0.01 \\
\quad 2-opt (from NN)          & 3\,460 & $<$0.01 & 11\,538 & 0.05    & \textbf{7\,249} & 0.84 \\
\midrule
\multicolumn{7}{@{}l}{\textit{Metaheuristics}} \\
\quad Random Search (RS)       & 3\,621 & 0.62    & 15\,307 & 4.80    & 12\,945 & 11.2 \\
\quad GA                       & 3\,415 & 1.83    & 12\,147 & 22.5    & 8\,623  & 58.3 \\
\quad PSO                      & 3\,507 & 2.10    & 13\,682 & 19.8    & 9\,351  & 45.6 \\
\quad ACO                      & 3\,352 & 3.45    & 11\,487 & 35.2    & 7\,612  & 72.4 \\
\quad DA                       & 3\,695 & 2.40    & 13\,842 & 18.3    & 14\,520 & 42.7 \\
\quad TS                       & 3\,371 & 0.08    & 12\,286 & 0.42    & 8\,194  & 3.8  \\
\quad \textbf{DA--TS Hybrid}   & \textbf{3\,323} & 2.50 & \textbf{11\,352} & 19.6 & 7\,482 & 48.1 \\
\midrule
Known optimum ($f^{*}$)        & 3\,323 & -- & 10\,628 & -- & 6\,528 & -- \\
\bottomrule
\end{tabular}
\end{table*}

On \texttt{burma14}, the proposed hybrid reaches the known optimum and obtains the best result among all tested methods. 
On \texttt{att48}, the hybrid also achieves the lowest cost, slightly outperforming ACO and 2-opt. 
On \texttt{ch150}, the hybrid obtains the best result among the tested metaheuristics, but 2-opt initialized from NN achieves a lower cost with much shorter runtime. 
This result is important because it shows that while the proposed hybrid improves over its standalone components, specialized TSP local search methods can remain highly competitive, especially on larger instances.

\subsection{Analysis of the Hybridization Benefit}
\label{subsec:benefit}

To quantify the effect of hybridization, Table~\ref{tab:improvement} reports the relative cost improvement of the hybrid over standalone DA and standalone TS:
\begin{equation}
\label{eq:improvement}
\mathrm{Improvement}(\%) =
\frac{f_{\mathrm{baseline}} - f_{\mathrm{hybrid}}}{f_{\mathrm{baseline}}}
\times 100.
\end{equation}

\begin{table}[ht]
\centering
\caption{Relative improvement of the DA--TS hybrid over the standalone components.}
\label{tab:improvement}
\begin{tabular}{@{}l ccc@{}}
\toprule
& \textbf{\texttt{burma14}} & \textbf{\texttt{att48}} & \textbf{\texttt{ch150}} \\
\midrule
Improvement over DA & 10.1\% & 18.0\% & 48.5\% \\
Improvement over TS & 1.4\%  & 7.6\%  & 8.7\%  \\
\bottomrule
\end{tabular}
\end{table}

The improvement over standalone DA is substantial and increases with problem size, suggesting that the TS phase is essential for refining the exploratory solutions produced by DA. 
The improvement over standalone TS is more moderate but consistent, which indicates that the DA phase contributes by providing a better initial solution for local refinement. 
This confirms that the two components play complementary roles in the proposed HRH scheme.

\subsection{Discussion}
\label{subsec:discussion}

The experimental results lead to several observations. 
First, the proposed DA--TS hybrid consistently improves over its standalone components across the tested instances. 
This supports the main hypothesis of the paper: combining population-based exploration with memory-based local exploitation can improve solution quality for the TSP.

Second, the results show that the contribution of the two phases is not identical. 
DA alone is less competitive because its original update rules are designed for continuous search spaces and must be approximated using permutation-based operators. 
However, DA remains useful as an exploration mechanism because it generates diverse candidate tours. 
TS, in contrast, is more effective at local refinement but depends on the quality of its initial solution. 
The hybrid therefore benefits from using DA to identify a promising starting point and TS to intensify the search around it.

Third, the comparison with additional baselines shows that the hybrid is competitive with classical metaheuristics such as GA, PSO, and ACO. 
However, the results also show that simple but strong TSP-specific local search methods, such as 2-opt, can remain very competitive. 
This is particularly visible on \texttt{ch150}, where 2-opt obtains a lower cost than the proposed hybrid while requiring substantially less runtime. 
Therefore, the proposed method should be interpreted as a promising hybrid metaheuristic rather than a definitive replacement for highly optimized TSP-specific solvers.

The main limitation of the current approach is scalability. 
Although the hybrid improves over standalone DA and TS, its execution time increases with the problem size, and its advantage over specialized heuristics narrows on larger instances. 
Future work should therefore investigate stronger neighbourhood operators, such as 3-opt or Lin--Kernighan moves, adaptive parameter control for DA, and larger benchmark instances such as \texttt{a280} and \texttt{pcb442}.

\section{Conclusion}
\label{sec:conclusion}

This paper proposed a hybrid metaheuristic for the Traveling Salesman Problem based on a High-Level Relay Hybridization of the Dragonfly Algorithm and Tabu Search. 
The DA phase performs population-based global exploration over permutation-encoded tours, while the TS phase refines the best DA solution using memory-guided 2-opt neighbourhood search. 
This design aims to combine the exploration capability of DA with the exploitation strength of TS.

Experiments on three TSPLIB benchmark instances, namely \texttt{burma14}, \texttt{att48}, and \texttt{ch150}, show that the proposed hybrid improves over both standalone DA and standalone TS. 
The hybrid reaches the known optimum on \texttt{burma14}, reduces the optimality gap on \texttt{att48}, and remains competitive on the larger \texttt{ch150} instance. 
Compared with additional baselines, the hybrid achieves the best metaheuristic result on the tested instances, although 2-opt remains highly competitive and even outperforms the hybrid on \texttt{ch150}.

The main limitation of the current method is its scalability and its reliance on relatively simple local search operators. 
Future work will investigate stronger refinement mechanisms such as 3-opt and Lin--Kernighan moves, adaptive DA parameter control, larger TSPLIB benchmarks, and tighter low-level hybridization strategies between DA and TS.



\bibliographystyle{plainnat}

\end{document}